\documentclass{article}
\usepackage{arxiv}
\usepackage[numbers,sort&compress]{natbib}
\usepackage[utf8]{inputenc} 
\usepackage[T1]{fontenc}    
\usepackage{ dsfont }
\usepackage{hyperref}       
\usepackage{url}
\usepackage{textcomp}
\usepackage{ gensymb }
\usepackage[ruled,vlined]{algorithm2e}
\usepackage{algorithmic}
\usepackage{booktabs}       
\usepackage{amsfonts}       
\usepackage{nicefrac}       
\usepackage{microtype}      
\usepackage{lipsum}
\usepackage{float}
\usepackage{graphicx}
\usepackage{multirow}
\usepackage{array}
\usepackage{comment}
\usepackage{longtable}
\usepackage{ragged2e}
\usepackage{subfig}

\title{Imbalanced Class Data Performance Evaluation and Improvement using Novel Generative Adversarial Network-based Approach: SSG and GBO}

\author{
  Md Manjurul Ahsan \\
  Department of Industrial and Systems Engineering\\
  University of Oklahoma\\
  Norman, Oklahoma-73071 \\
  \texttt{ahsan@ou.edu} \\
   \And
 Md Shahin Ali \\
  Department of Biomedical Engineering\\
  Islamic University\\
  Kushtia, 7003, Bangladesh\\
  \texttt{shahin@std.iu.ac.bd}\\
  \And
 Zahed Siddique \\
  School of Aerospace and Mechanical Engineering\\
  University of Oklahoma\\
  Norman, Oklahoma-73019\\
  \texttt{zsiddique@ou.edu}} 

\begin{document}
\maketitle
\begin{abstract}
Class imbalance in a dataset is one of the major challenges that can significantly impact the performance of machine learning models resulting in biased predictions. Numerous techniques have been proposed to address class imbalanced problems, including, but not limited to, Oversampling, Undersampling, and cost-sensitive approaches. Due to its ability to generate synthetic data, oversampling techniques such as the Synthetic Minority Oversampling Technique (SMOTE) is among the most widely used methodology by researchers. However, one of SMOTE's potential disadvantages is that newly created minor samples may overlap with major samples. As an effect, the probability of ML models' biased performance towards major classes increases. Recently, generative adversarial network (GAN) has garnered much attention due to its ability to create almost real samples. However, GAN is hard to train even though it has much potential. This study proposes two novel techniques: GAN-based Oversampling (GBO) and Support Vector Machine-SMOTE-GAN (SSG) to overcome the limitations of the existing oversampling approaches. The preliminary computational result shows that SSG and GBO performed better on the expanded imbalanced eight benchmark datasets than the original SMOTE. The study also revealed that the minor sample generated by SSG demonstrates Gaussian distributions, which is often difficult to achieve using original SMOTE.

\end{abstract}
\maketitle
\section{Introduction}
\begin{figure}[!ht]
    \centering
    \includegraphics[width=.5\textwidth]{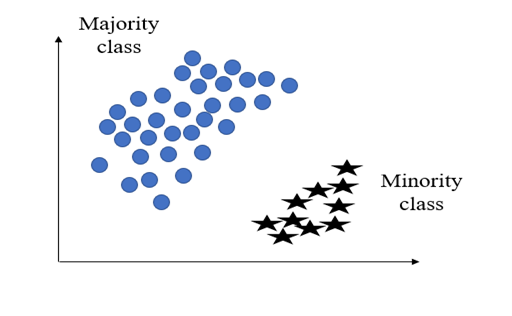}
    \caption{Hypothetical example of majority and minority class}
    \label{fig:fig1}
\end{figure}

An imbalanced class ratio with datasets is a potential challenge in machine learning (ML)-based model development systems~\cite{yang2020rethinking}. Class imbalance occurs when the total number of samples from one class is significantly higher than the other classes~\cite{fernandez2011addressing}. 
In both binary and multiclass classification situations, this inequality can be observed~\cite{wang2021research}. The data class with the lowest sample calls the minor class, and the data class with the highest sample calls the major class~\cite{wang2021research}. Major class frequently refers to a negative class in binary classification problems, whereas minor class refers to a positive class. It is currently a significant issue in various domains such as biology, health, finance, telecommunications, and disease diagnosis~\cite{ahsan2022machine}. As an effect, it is considered one of the most severe problems in data mining~\cite{yang200610}. Figure~\ref{fig:fig1} depicts a two-dimensional representation of the major and minor classes.

Most of the ML algorithms are built in such a way that they perform well with balanced data, but are unable to perform on the imbalanced dataset~\cite{garcia2009evolutionary}. Therefore, several conditions, such as detecting credit card fraud or identifying malignant tumor cells, are difficult to accomplish using typical ML algorithms, where the primary goal is to identify the positive samples (rare samples)~\cite{das2013handling}.
A well-known example is Caruana et al. (2015)'s study, that sought to determine
which pneumonia patients might be hospitalized and which might be discharged home~\cite{caruana2015intelligible}. Unfortunately, their proposed approach generated misleading results for patients with asthma or chest pain by estimating a lower likelihood of dying.

Numerous ML algorithms have been proposed, and their performance remains biased toward the major class. For instance, consider an imbalanced dataset comprising 10 924 non-cancerous (majority class) and 260 malignant (minority class) cell image data. Using traditional ML algorithms, there is a greater chance that the classification will exhibit a 100\% accuracy for the major class and 0\% -10\% accuracy for the minor class, resulting in the probability of classifying 234 minor class as the major class~\cite{fletcher2021addressing}. Therefore, 234 patients with cancer would be misdiagnosed as non-cancerous. Such an error is more costly in medical treatment, and a misdiagnosis of a malignant cell has significant health repercussions and may result in a patient’s death~\cite{longadge2013class}. 

\begin{figure*}[!ht]
    \centering
    \includegraphics[width=\textwidth]{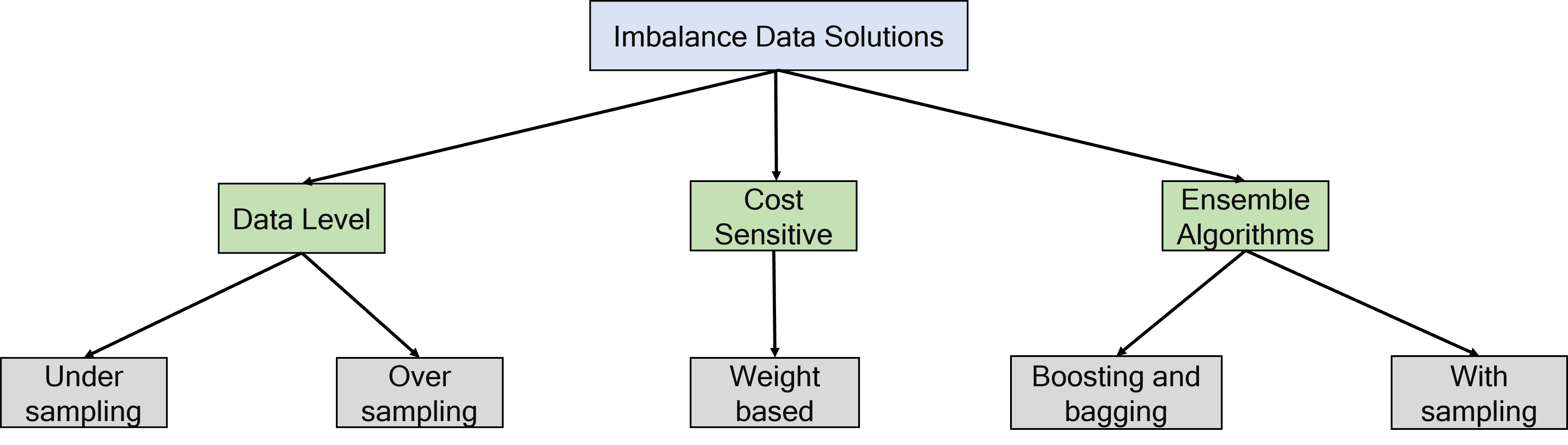}
    \caption{Major approaches to handle CIP in the machine learning domain.}
    \label{fig:isl}
\end{figure*}

Algorithms and techniques to address class imbalanced problems (CIP) are usually classified into three main categories: data level, cost-sensitive, and ensemble algorithms (as shown in Figure~\ref{fig:isl}). 

In data level solutions Oversampling approaches are mostly used where minor class data is oversampled by applying different techniques.
Oversampling techniques that are often employed include Adaptive Synthetic (ADASYN), Random Oversampling, Synthetic Minority Oversampling Techniques
(SMOTE), and Borderline SMOTE~\cite{kaur2019systematic}. Among all Oversampling approaches, Chawla’s SMOTE is the most popular and commonly utilized~\cite{wang2021research}. However, traditional SMOTE produces more noise and is unsuitable for high-dimensional data. To resolve these issues, Wang et al. (2021) proposed active learning-based SMOTE~\cite{wang2021research}. Zhang et al. (2022) proposed SMOTE-RkNN (reverse k-Nearest Neighbors), a hybrid oversampling technique to identify the noise instead of local neighborhood information~\cite{zhang2022smote}. Maldonado et al. (2022) demonstrated that traditional SMOTE faces major difficulties when it comes to defining the neighborhood to generate additional minority samples. To overcome these concerns, the authors proposed a feature-weighted oversampling, also known as (FW-SMOTE)~\cite{maldonado2022fw}. Aside from that, SMOTE is often computationally costly, considering the time and memory usage for high-dimensional data. Berando et al. (2022) pioneered the use of C-SMOTE to address time complexity issues in binary classification problems~\cite{bernardo2022extensive}. Obiedat et al. (2022) presented SVM-SMOTE combined with particle swarm optimization (PSO) for sentiment analysis of customer evaluations; however, the proposed algorithms remained sensitive to multidimensional data~\cite{obiedat2022sentiment}

On the other hand, Undersampling procedures reduce the sample size of the major classes to create a balanced dataset. Near-miss Undersampling, Condensed Nearest Neighbour, and Tomek Links are three popular Undersampling methods, that are used more frequently~\cite{kaur2019systematic}. Once the data sample is reduced using Under-sampling techniques, there is a higher chance that it will also eliminate many crucial information from the major class. As a result, Oversampling is generally preferred rather than Undersampling by researchers and practitioners~\cite{kotsiantis2006handling}.

Most ML algorithms consider that all the misclassification performed by the model is equivalent, which is a frequently unusual case for CIP, wherein misclassifying a positive (minor) class is considered as the worse scenario than misclassifying the negative (major) class. Therefore, in the cost-sensitive approach higher penalty is introduced to the model for misclassifying the minor samples. In this process, the cost is assigned based on the error made by the model. Suppose, if the algorithm fails to classify the minor class, then the penalty will be higher (i.e., 10 for each misclassification), and if the algorithm fails to classify the major class, then the penalty will be lower (i.e., 1 for each misclassification). Shon et al. (2020) proposed hybrid deep learning based cost-sensitive approaches to classify kidney cancer~\cite{shon2020classification}. Wang et al. (2020) used multiple kernel learning-based cost-sensitive approaches to generate synthetic instances and train the classifier simultaneously using the same feature space~\cite{wang2020multiple}. One of the potential drawbacks of the cost-sensitive approach is that no defined protocol can be used to set the penalty for misclassification. Therefore, adjusting weight is less preferred due to its complexity of use. The cost (weight) for misclassification is set by the expert’s opinions or by manually experimenting until the appropriate cost is identified, which is very time-consuming. Further, determining the penalty requires measuring the impact of the features and considering various criteria. However, such a procedure becomes more complex with multidimensional and multiclass label data.
 
Several algorithm-based solutions have been proposed to improve the effect of ML classification on the imbalanced dataset. Galar et al. (2013) suggested an ensemble-based solution (EUSBoost), which integrated random Undersampling and boosting algorithms. The author assert that their proposed approach can resolve the imbalanced class problem’s overfitting issues~\cite{galar2013eusboost}. Shi et al. (2022) proposed an ensemble resampling based approach considering sample concatenation (ENRe-SC). According to the author, the proposed strategy can mitigate the adverse effect of removing the major class caused by Undersampling approaches~\cite{shi2022resampling}. Muhammad et al. (2021) proposed an evolving SVM decision function that employs a genetic method to tackle class imbalanced situations~\cite{mohammed2021stacking}. Jiang et al. (2019) proposed generative adversarial network (GAN) based approaches to handle the imbalanced class problem in time series data~\cite{jiang2019gan}. Majid et al. (2014) employed K-nearest neighbor (KNN) and support vector machines (SVM) to detect human breast and colon cancer. The authors use a two-step process to address the imbalance problem: a preprocessor and a predictor. On one hand, Mega-trend diffusion (MTD) is used in the preprocessing stage to increase the minority sample and balance the dataset. On the other hand, K-nearest neighbors (KNN) and SVM are used in the predictor stage to create hybrid techniques MTD-SVM and MTD-KNN. Their study result shows that MTD-SVM outperformed all other proposed techniques by demonstrating an accuracy of 96.71\%~\cite{majid2014prediction}. Xiao et al. (2021) introduced a deep learning (DL)-based method known as the Wasserstein generative adversarial network (WGAN) model and applied it to three different datasets: lung, stomach, and breast cancer. WGAN can generate new instances from the minor class and solve the CIP ratio problem~\cite{xiao2021cancer}.

GAN has become a widely utilized technique in computer vision domains. GAN’s capacity to generate real images from random noise is one of its potential benefits~\cite{lin2018pacgan}. This dynamic characteristic contributes to GAN’s appeal, as it has been used in nearly any data format (i.e., time-series data, audio data, image data)~\cite{frid2018synthetic}. Sharma et al. (2022) showed that using GAN, it is possible to generate data in which the sample demonstrates better Gaussian distribution, which is often difficult to achieve using traditional imbalanced approaches. Their proposed GAN-based approaches show comparatively 10\% higher performance than any other existing techniques while producing minor samples that are almost real~\cite{sharma2022smotified}. However, one major drawback of their suggested approach is that the model is hardly stable and very time-consuming in generating new samples. Therefore, an updated stable GAN-based oversampling technique might play a crucial role in tackling class imbalanced problems. Considering this opportunity, in this work we have presented an updated GAN-based oversampling technique. Our technical contributions can be summarized as follows:
\begin{enumerate}
    \item Taking into account the advantages of two algorithms: SVM-SMOTE and GAN, we proposed two Oversampling strategies: GAN-based Oversampling (GBO) and SVM-SMOTE-GAN (SSG).
    \item On eight benchmark datasets, oversampling and non-oversampling approaches are evaluated, and our proposed model suppresses the previous research outcome in terms of different imbalanced datasets.
    \item The proposed SSG and GBO can produce different data samples with gaussian distributions, which was not addressed in the prior study.
\end{enumerate}

Considering these possibilities, this study presents two GAN-based Oversampling strategies: GAN-based Oversampling (GBO) and SVM-SMOTE-GAN (SSG). The experimental results reveal that the neural network classification on expanded data by the proposed GBO and SSG algorithm performs better than the traditional SMOTE on the imbalanced datasets used in this study.

 \section{Methods}
 \subsection{SVM-SMOTE Algorithm}
\begin{figure*}
    \centering
    \includegraphics[width=.8\textwidth]{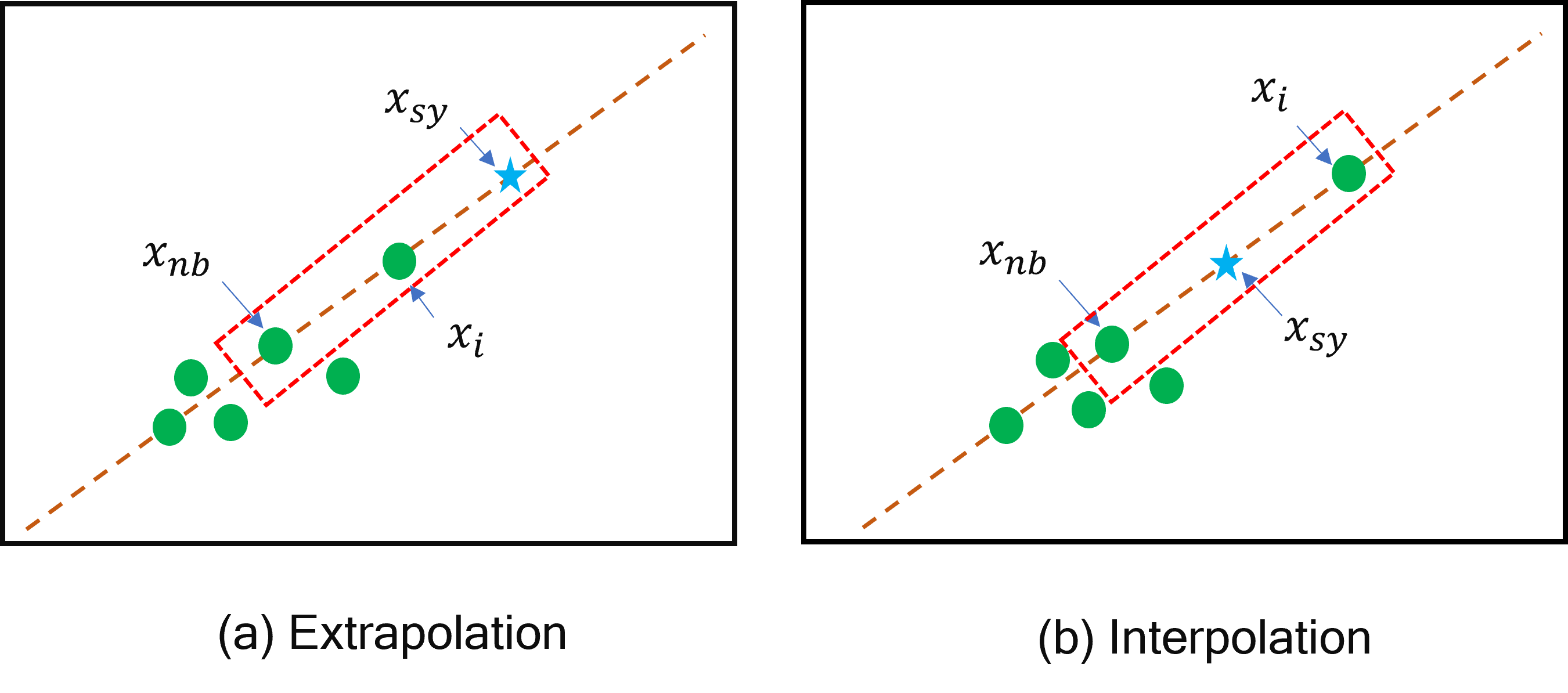}
    \caption{Process of generating synthetic sample using (a) extrapolation and (b) interpolation.}
    \label{fig:svmsmote}
\end{figure*}

 SVM-SMOTE is one of the several variations of SMOTE algorithms that have been introduced over the years. It creates new observations from the sample of the minority class that are harder to classify~\cite{tang2008svms}. SVM-SMOTE uses instances and observations from the minor class that is the support vectors of a support vector machine. There are two methods to create the data. If the support vector is surrounded by neighbors from the minority class, it will create the data by extrapolation. However, if the support vector is surrounded mainly by neighbors from the majority class, it will create the data by interpolation. The following is a detailed description of the SVM-SMOTE process~\cite{tang2008svms}.
 \begin{itemize}
     \item Step 1: SVM is trained on the entire dataset to find the support vectors from the minor and the major class. SVM constructs hyperplanes with the highest distance to the closest training points (the margins) to seperate classes. The sample on the margin (as shown in Figure~\ref{fig:svmsmote}) are the support vectors.
     \item Step 2a: Combined with the nearest neighbor algorithm is trained on the entire dataset (minority and majority classes together). Then it will look at the closest neighbors of each support vector on the minor class. If most of the neighbors from the support vectors belong to the minority class, then it will perform extrapolation; in other words, it
will expand the boundaries by creating a sample out of the boundary.


     In the case of extrapolation, In Figure~~\ref{fig:svmsmote}(a), if line $S_v$ is the support vector line and  $X_{nb}$ is the nearest neighbors value of instance $x_i$, then the artificial instance generated by SVM-SMOTE is $X_{sy}$. 
     
     \item Step 2b: If most of the neighbors of the support vector belong to the majority class, then the algorithm will perform the interpolation. 
In the case of interpolation, (refer to Figure~\ref{fig:svmsmote}(b)), if line $S_v$ is the support vector line and  $X_{nb}$ is the nearest neighbors value of instance $x_i$, then the artificial instance generated by SVM-SMOTE is $X_{sy}$. The idea is that the new example will fall within the two existing observations from the minority. Therefore, the SVM-SMOTE will not expand the boundaries of the minority class; instead, it will create samples within the boundary~\cite{brownlee2020imbalanced}.
 \end{itemize}
A pseudocode of the SVM-SMOTE based approach is presented in Algorithm~\ref{svmsmote}~\cite{kumar2022addressing}.
\begin{algorithm}
\caption{Pseudocode for SVM-SMOTE}\label{svmsmote}
\begin{algorithmic}
\STATE // Input: X Training sample, N sampling level (100, 200,.......\%), K nearest neighbors, m number of nearest neighbor to decide sampling type, $SV^+$ set of positive support vectors (SVs), T number of artificial instances to be created, ammount array contains the number of artificial instances, nn array contains k positive nearest neighbors of each positive SV.

// initialize parameters
\STATE T $\gets (N/100) \times$ \textbar $X$\textbar

\STATE 1:  Compute $SV^+$ by training SVMs with X
\STATE 2: Compute amount by evenly distributing $T$ among $SV^+$

\STATE 3: Compute $nn$
\STATE 4: For each $sv^+_{i} \in SV^+$, compute $m$ nearest neighbors on $X$.
\STATE 5: Create $ammount[i]$ artificial positive instances.
\STATE $X_new = X \cup {X^+_{new}} $
\STATE 6: Output : $X_{new}$ : New oversampled instances
\end{algorithmic}
\end{algorithm}
\subsection{Motivation}
Major and minor classes overlapped due to the oversampling, which is one of the potential drawbacks of applying SMOTE as oversampling techniques~\cite{wang2021research}. The SVM-SMOTE algorithm is used to reduce the marginalization, which creates the hyperplane between major and minor classes. However, SVM is sensitive to imbalanced data by nature, which potentially often affects data distribution~\cite{brownlee2020imbalanced}. Therefore, GAN-based two Oversampling approaches are proposed--GBO and SSG-- to reduce the marginalization and improve the data distribution between major and minor samples.
\subsubsection{Modified GAN}
GAN is conventionally designed to create images that look almost similar to the real images; therefore, it is not ideal for oversampling approaches in other data types (i.e., numerical, categorical, and text data)~\cite{sharma2022smotified}. However, since GAN has the powerful ability to augment the images, this concept can be used to create a new but real sample which can also be used as an oversampling approach in CIP-related problems. Since original GAN requires a good amount of data, a lack of data from the minor sample may not be helpful to create enough samples using GAN. However, using cross-validation techniques, these problems can be solved. GAN has two neural networks where the generator's goal is to generate fake samples that confuse the discriminator to classify as "real." To maximize its performance it is necessary to optimize the discriminator's loss when data comes from the generator to force the discriminator to classify the fake samples as real. Contrarily, the goal of the discriminator is to distinguish between original and fake data by minimizing the loss
when given batches of both original and generated fake data samples~\cite{creswell2018generative}. The discriminator's loss can be calculated as follows~\cite{goodfellow2014generative}:
\begin{equation}
\centering
    \max_D\mathds{E}_x[logD(x)]+\mathds{E}_z[log(1-D(G(z)))] 
\end{equation}
 Here, $D(x)$ denotes the probability output of real data, $x$ of the discriminator, and $D(G(z))$ denotes the probability output of generated samples by $z$. The generator loss can be calculated as follows~\cite{goodfellow2014generative}:
 \begin{equation}
     \min_G-\mathds{E}_z[logD(G(z))]
 \end{equation}
 A pseudocode of the GAN based approach is presented in Algorithm~\ref{gan}.
\begin{algorithm}
\caption{Pseudocode for GAN}\label{gan}
\begin{algorithmic}[5]
\STATE // Input: data sample x
and noise samples z (generated randomly).

// initialize parameters

// $m_i$ is minibatch indices for $i^{th}$. $n_{fake}$ denotes to numbe of fake samples needed.
index and $T$ is total iterations.
\STATE GAN ($x,z, n_{fake}$)

\FOR{t=1:T} 

\STATE // step size $S$ = 1

\STATE // subscript $d$ and $g$ refers to
discriminator and generator
\FOR{$s=1:S$}
\STATE $g_d \gets$
\STATE $SGD(- \log D(x) - \log(1 - D(G(z)), W_D , m_i)$
\STATE $W_d \gets weights(g_d , W_d )$
\STATE $W_g \gets weights(g_g, W_g)$

\ENDFOR
\ENDFOR
\STATE $x' \gets$ FinalOutput ($Model_d(W_d,x,z), 
Model_g (W_g, x,z), n_{fake})$
\RETURN $x'$
\end{algorithmic}
\end{algorithm}
 
 The proposed SSG is developed by modifying and combining SVM-SMOTE and GAN. Here, only the random sample of GAN is replaced with the collection of Oversample minority instances from SVM-SMOTE. The updated Discriminators loss can be expressed as follows:\\
\begin{equation}
   \max_D\mathds{E}_{x^*}[logD(x^*|x)]+\mathds{E}_u[log(1-D(G(u)))] 
\end{equation}
and the generator loss can be express as:\\
\begin{equation}
     \min_G\mathds{E}_z[logD(G(u))]
\end{equation}
Where $x^*$ is the training sample of the minor class, and $u$ is the Oversampled data of the same classes generated by the SVM-SMOTE.

\begin{figure*}
    \centering
    \includegraphics[width=\textwidth]{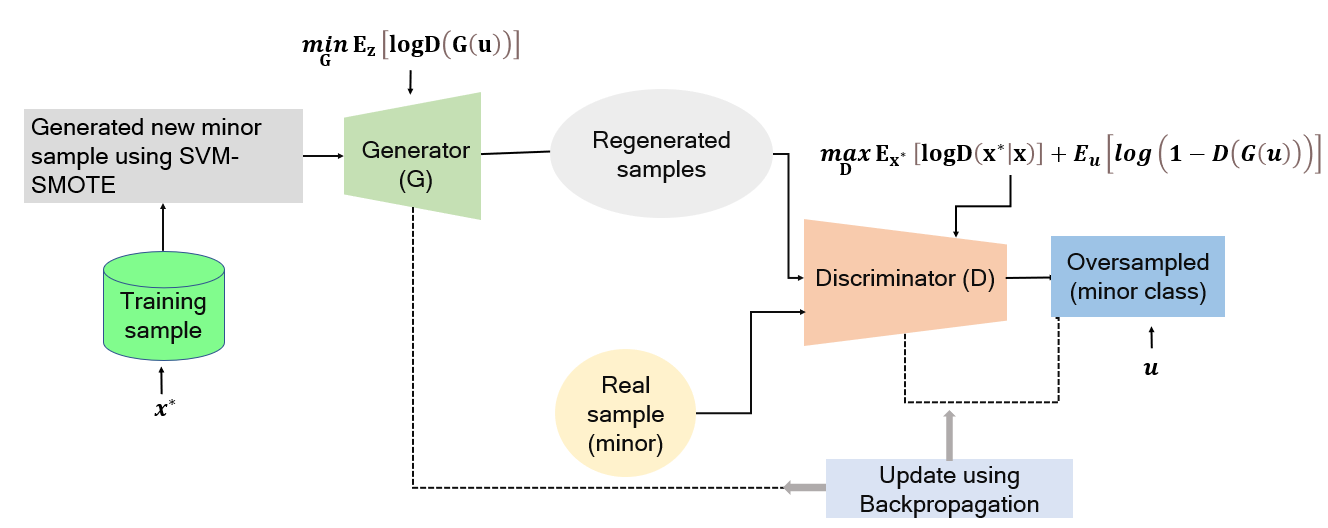}
    \caption{Process of generating “fake” samples using SSG.}
    \label{fig:SVMgan}
\end{figure*}

As illustrated in Figure~\ref{fig:SVMgan}, SSG is composed of two sections. The first section replaces the random samples with the collection of oversamples from the SVM-SMOTE. The second section continues the GAN process by incorporating the new samples from the SVM-SMOTE. The key difference between the proposed SSG and GBO is using ready-made samples generated by SVM-SMOTE instead of arbitrary samples. As an effect, it helps produce better over-samples by providing better input samples. Using the inherent synergy of SVM-SMOTE and GAN, the naive GAN is guided to get a prompt start using "realistic" data even before additional tuning is performed. A pseudocode of the proposed GAN-based approach is presented in Algorithm~\ref{modify}.
\begin{algorithm}
\caption{: Pseudocode for Proposed modified GAN based approaches}\label{modify}
Step 1 $\rightarrow$ Input: minor samples $X^*$ from the training sample $x$ of size $N$ with $N$ -- $n$ over-samples;\\
Step 2 $\rightarrow$ parameter $k$ for KNN to find the nearest neighbor\\
Step 3 $\rightarrow$ Execute SVM-SMOTE given in Algorithm~\ref{svmsmote} and then GAN given in Algorithm~\ref{gan}\\
1 $u \leftarrow$ call Algorithm~\ref{svmsmote} ($x^*,k$)// generate over-sampled minor data $u$.\\
2 $u \leftarrow$ call Algorithm~\ref{gan} ($x^*$, $u$, $N$ - $n$).\\
\end{algorithm}
\section{Classification methods and evaluation index}
\begin{figure*}
    \centering
    \includegraphics{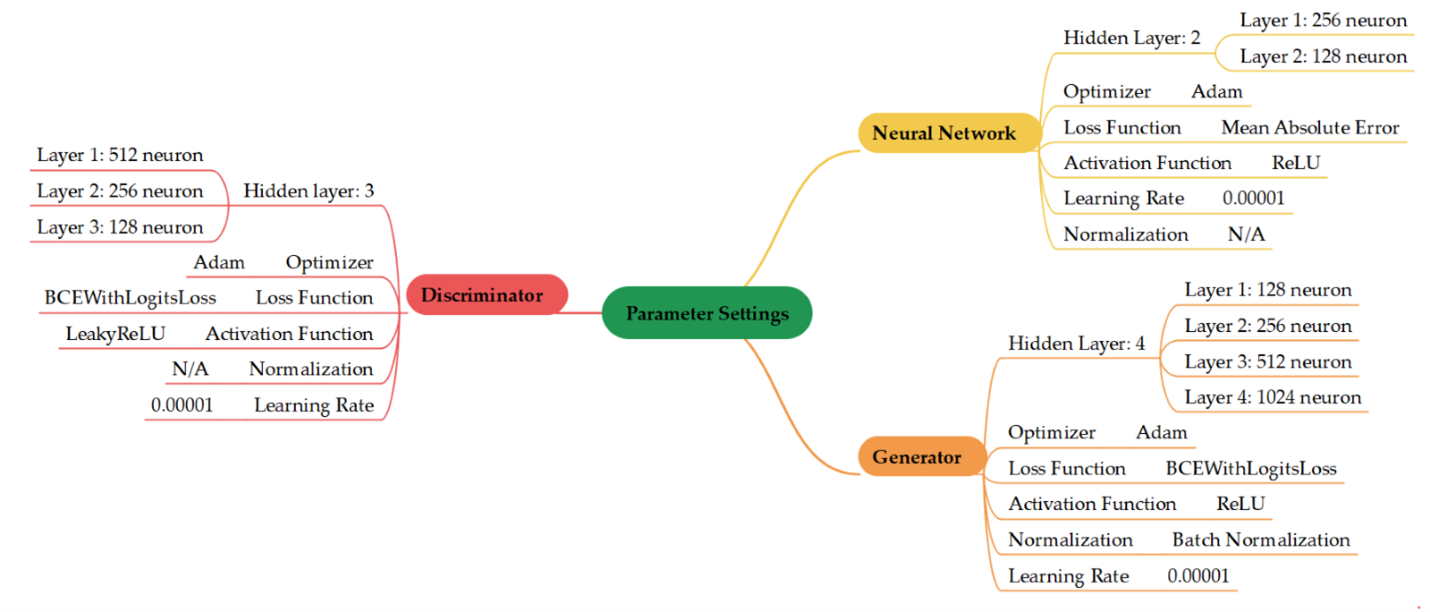}
    \caption{Parameter settings used in this study.}
    \label{fig:para}
\end{figure*}

With advancements in computational power and machine learning (ML) tools, neural network (NN)-based approaches are now most widely used on small and large datasets due to their robust and higher accuracy than other existing ML algorithms~\cite{pham2022identifying,ahsan2020deep}. One of the potential advantages of using the NN approach is that, NN based methods are less sensitive compared to traditional ML algorithms (i.e., KNN, support vector, logistic regression)~\cite{islam2022improving,ahsan2022machine}. The proposed NN approach was developed based on tuning the following parameters: batch size, learning rate, epochs, number of hidden layers, and optimization algorithms followed followed by~\ref{danala2022comparison,islam2022improving1}. The parameter settings for proposed NN are illustrated in Figure~\ref{fig:para}.

To carry out the whole experiment procedure following steps are used as follows:
\begin{enumerate}
      \item Initially, the dataset was divided into two sets, where 80\% of the data belonged to the train set, and 20\% of the data belonged to the test set. 
      \item Dataset was fit to different algorithms such as SMOTE, GBO, and SSG to create Oversample data separately and save it as a new dataset.
      \item The newly created balanced dataset was trained and tested with the designed NN models.
      \item The simulation was run ten times, and the result was presented by averaging all the results.
\end{enumerate}
\subsection{Experimental evaluation index}
The results are presented regarding the accuracy, precision, recall, and F1 score with standard deviation. Suppose the dataset is classified into two types: positive and negative samples. Then the evaluation matrix is as follows~\cite{ahsan2020deep}:
\begin{equation}
Accuracy=\frac{T_p+T_n}{T_p+T_n+F_p+F_n}
\end{equation}
\begin{equation}
  Precision=\frac{T_p}{T_p+F_p}  
\end{equation}
\begin{equation}
   Recall=\frac{T_p}{T_n+F_p} 
\end{equation}
\begin{equation}
  F1=2\times\frac{Precision\times Recall}{Precision+Recall}  
\end{equation}
Where,
True Positive ($T_p$)= Positive sample classified as Positive

False Positive ($F_p$)= Negative samples classified as Positive 

True Negative ($T_n$)= Negative samples classified as Negative

False Negative ($F_n$)= Positive samples classified as Negative.

\section{Simulation experiment}
\subsection{Experimental environment. }
The experiment was carried out using eight benchmark datasets obtained from the open-source UCI (University of California Irvine) repository: page blocks, Ecoli, Winequality, Abalone, Ionosphere Spambase, Shuttle, and Yeast dataset. Detailed information regarding those datasets is summarized in Table~\ref{tab:tab1}~\cite{asuncion2007uci}.
\begin{table*}[!ht]
\caption{Properties of Imbalanced dataset utilized in this study.}
    \centering\resizebox{\textwidth}{!}{
    \begin{tabular}{ccccccc}\toprule
         Dataset &	Total sample& 	Minor class& 	Major class &	Total features& 	Minority class(\%)& Description\\\midrule
Page-blocks &	471& 	28& 	443& 	10& 	5.94& Classify blocks from page layout\\
Ecoli& 	335& 	20& 	315& 	7& 	5.97& Protein localization\\ 
Winequality& 	655& 	18& 	637& 	10& 	2.74& Classify the wine quality\\ 
Abalone& 	4177& 	840& 	3337& 	8& 	20.1& Predict the age of abalone\\ 
Ionosphere& 	351& 	126& 	225& 	34& 	35.71& Classify radar returns\\ 
Spambase& 	4601& 	1812& 	2788& 	57& 	39.39 & Classify spam and non-spam email\\	
Shuttle& 	58000& 	170& 	57830& 	9& 	0.294& 80\% sample belongs to major class\\ 	 
Yeast&		513&		51&		462&	8&		9.94& Predicting protein localization cite.\\\bottomrule
    \end{tabular}}
    
    \label{tab:tab1}
\end{table*}

The proposed oversampling models were implemented and experimented with using the Anaconda modules with Python 3.8 on a traditional laptop with typical configurations (Windows 10, 16 GB of RAM, and Intel Core I7-7500U). The experiment was performed ten times and the final mean, maximum, and standard deviation values were used to determine the model's overall performance. An early stope method is utilized to avoid overfitting. For validation purposes, the dataset was split into the following training set/testing set ratios: 75:25, 70:30, and 80:20. Split ratios of this type are frequently employed in machine learning approaches to evaluate and validate models. The optimal training and testing accuracy results were obtained when the dataset was randomly divided into an 80\% training set and a 20\% testing set~\cite{ahsan2020deep}.
\subsection{Numerical experiment}
The experiment was carried out with eight different datasets (as shown in Table~\ref{tab:tab1}), considering without sampling the data and oversampling the data using three different methods— SMOTE, GBO, and SSG. Then, to conduct the appropriate comparison, a neural network (NN) is used to classify the expanded and original data. As previously stated, each experiment was repeated ten times to ensure scientifically valid and appropriate experimental results. The GBO model and SSG are tested with different parameter settings. Due to the significant effort required to tune those parameters manually, only three primary parameters were optimized: batch size, epochs, and learning rate. Grid search is used to determine the best parameters~\cite{ahsan2020deep}. The following parameters were used for grid search methods:\\
\begin{center}
   Batch size = [32, 64,128]\\
Number of epochs = [40, 50, 100, 200, 500]\\
Learning rate = [0.00001, 0.0001, 0.001, 0.01]\\  
\end{center}

\begin{figure*}
    \centering
    \includegraphics[width=\textwidth]{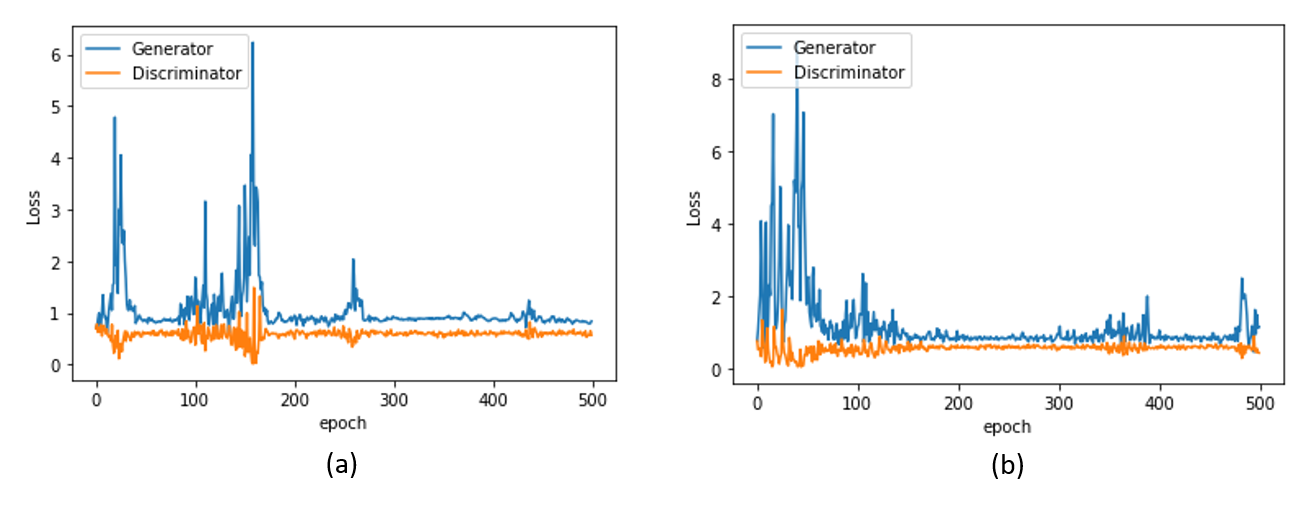}
    \caption{Loss per epoch during the training with (a) GBO and (b) SSG.}
    \label{fig:dgloss}
\end{figure*}

Figure~\ref{fig:SVMgan} illustrates the optimal parameters obtained using grid search methods for discriminator, generator, and neural networks. The generator network contains four hidden layers with 128, 256, 512, and 1024 neuron respectively. The discriminator network has three hidden layers with 512, 256, and 128 neurons. The training procedure is carried out with the Adam optimizer and a binary cross-entropy function with a training data batch size of 32 and an initial learning rate of 0.00001.\\
Figure~\ref{fig:dgloss} illustrates the loss per epoch on Abalone datasets for the generator and discriminator during the training. From the Figure, it is clear that using a minor sample generated by SVM-SMOTE shows more stability (refer to Figure~\ref{fig:dgloss}b) than the sample created from random noise (Figure~\ref{fig:dgloss}a).
\section{Results and Discussion}
The overall performance of eight benchmark datasets without and with oversampling approaches (SMOTE, GBO, SSG) was measured on the training and testing set. The model's performance in terms of accuracy for the training set and accuracy, precision, recall, and F1 score for the testing set are measured and presented in Table~\ref{tab:result}, where bold font specify the better experiment results. From Table~\ref{tab:result}, it is clear that the proposed SSG outperformed all other techniques on all datasets across all measures, while the proposed GBO becomes the second-best algorithm except on the Page blocks dataset. On Page blocks SMOTE showed best result on the train set by achieving an accuracy of 99.51\%. On Yeast dataset, GBO showed the highest accuracy on train set (96.07\%), where SSG demonstrated best performance across all measures on the test set. 
\begin{table*}[!ht]
\caption{Performance evaluation of different Oversampling techniques used in this study on eight highly imbalanced benchmarks datasets. $M_e$--Mean; $M_a$--Maximum; $S_d$--Standard deviation.}
    \centering
    \resizebox{\textwidth}{!}{\begin{tabular}{ccccccc}\toprule
         Dataset&	Sampling strategy&	Train&\multicolumn{4}{c}{Test}\\\cmidrule{4-7}
         &&accuracy ($M_e$,$M_a$,$S_d$)&accuracy ($M_e$,$M_a$,$S_d$)& precision ($M_e$,$M_a$,$S_d$)&recall ($M_e$,$M_a$,$S_d$) & f1 score ($M_e$,$M_a$,$S_d$)\\\midrule
         
\multirow{4}{*}{Page blocks}& Without-oversampling&	96.27\% (96.27\%, 0.0000)&	97.68\% (98.94\%, 0.6657)&	0.9741 (0.9890, 0.0082)&	0.9774 (0.9895, 0.0063)&	0.9768 (0.9895, 0.0066)\\
&SMOTE&\textbf{99.51\%} (99.71\%, 0.1200)&	98.21\% (98.95\%, 0.8666)&	0.9832 (0.9898, 0.0076)&	0.9821 (0.9895, 0.0086)&	0.9821 (0.9895, 0.0086)\\
         &GBO&	97.96\% (98.02\%, 0.0729)&	98.53\% (98.95\%, 0.5436)&	0.9844 (0.9890, 0.0060)&	0.9854 (0.9896, 0.0053)&	0.9853 (0.9895, 0.0054)\\
         &SSG&	98.02\% (98.02\%, 0.0000)&	\textbf{98.84\%} (98.94\%, 0.3328)&	\textbf{0.9878} (0.9890, 0.0038)&\textbf{0.9885} (0.9896, 0.0032)&\textbf	{0.9884} (0.9895, 0.0033)\\\hline

        \multirow{4}{*} {Ecoli}&	Without-oversampling&	93.29\% (93.38\%, 0.0316)&	95.01\% (95.01\%, 0.0000)&	0.9732 (0.9732, 0.0000)&	0.9601 (0.9601, 0.0000)&	1.0 (1.0, 0.0000)\\
         &SMOTE&	98.70\% (99.40\%, 0.5011)&	95.82\% (97.01\%, 0.6293)&	0.9664 (0.9748, 0.0044)&	0.9622 (0.9622, 0.00)&	0.9582 (0.9701, 0.0062)\\
         &GBO&	92.87\% (95.59\%, 1.8189)&	88.65\% (91.04\%, 2.2470)&	0.9202 (0.9351, 0.0140)&	0.9407 (0.9407,0.00)&	0.8865 (0.91044, 0.0224)\\
         &SSG&\textbf{97.87\%} (98\%, 0.1032)&\textbf{97.02\%} (97.11\%, 0.0316)& \textbf{0.9848} (0.9848, 0.00)&\textbf{0.9702} (0.9702, 0.00)&\textbf{1.00} (1.00, 0.00)\\\hline
                 \multirow{4}{*}{Winequality}&	Without-oversampling&	97.5\% (97.77\%, 0.63)&	95.21\% (95.41\%, 0.63)&	0.94 (0.94, 0.0000)&	0.9220 (0.9220, 0.0000)&	.9230 (0.0000, 0000)\\
	&SMOTE&	98.25\% (98.59\%, 0.3100)&	91.15\% (91.60\%, 0.39)&	0.9160 (0.9230, 0.0049)&	0.9250 (0.9330, 0.005)&	0.9083 (0.9240, 0.007)\\
	&GBO&	98.32\% (98.47\%, 0.1300)&	95.49\% (96.19\%, 0.2500)&	0.9270 (0.9280, 0.0012)&	0.9191 (0.9101, 0.0001)&	0.9450 (0.9460, 0.0024)\\
	&SSG&\textbf{98.43\%} (98.47\%, 0.0620)&\textbf{95.42\%} (95.52\%, 0.0300)& \textbf{0.9420} (0.9760, 0.0240)&\textbf{0.9233} (0.9540, 0.0210)&
	\textbf{0.9620} (1.0, 0.0264)\\\hline
\multirow{4}{*}{Yeast}&	Without-oversampling&	90.87\%(91.95\%, 0.4177)&	89.42\% (92.23\%, 1.0680)&	0.8738 (0.9381, 0.0377)&	0.9028 (0.9286, 0.0126)&	0.9174 (1.0, 0.0444)\\
&	SMOTE&	92.02\% (92.18\%, 0.1530)&	90.67\% (92.23\%, 1.532)&	0.9142 (0.9270, 0.0124)&	0.9294 (0.9360, 0.0061)&	0.9067 (0.9223, 0.0444)\\

&	GBO&\textbf{96.07\%} (96.63\%, 0.42)&	92.91\% (94.17\%, 0.7992)&	0.9146 (0.9329, 0.01176)&	0.9344 (0.9453, 0.0068)&	0.9291 (0.9417, 0.0079)\\
	&SSG&	95.91\% (97.304\%, 1.115)&\textbf{94.17\%} (95.15\%, 0.7929)&\textbf{0.9388} (0.9483, 0.0074)&\textbf{0.9396} (0.9494, 0.0080)&\textbf{0.9417} (0.9514, 0.0079)\\\hline

\multirow{4}{*}	{Abalone}&	Without-oversampling&	90.43\% (90.66\%, 0.3450)	&90.16\% (90.55\%, 0.3601)&	0.8996 (0.9035, 0.0050)&
	0.8991 (0.9026, 0.0043)&	0.9016 (0.9055,0.0036)\\
	&SMOTE&	88.40\% (88.74\%, 0.2452)&	83.95\% (84.80\%, 0.8725)&	0.8526 (0.8597, 0.0073)&	0.8964 (0.8989, 0.0016)&
	0.8395 (0.8480, 0.0087)\\
	&GBO&	93.72\% (93.94\%, 0.2240)&	89.98\% (90.43\%, 0.2328)&	0.9001 (0.9038, 0.0023)&	0.9008 (0.9049, 0.0028)&	0.9008 (0.9049, 0.0028)\\
	&SSG&\textbf{94.11\%} (94.24\%, 0.1012)&\textbf{90.25\%} (90.43\%, 0.2342)&
	\textbf{0.9014} (0.9039, 0.0018)&\textbf{0.9012} (0.9047, 0.0018)&\textbf{red}{0.9025} (0.9043, 0.0023)\\\hline
\multirow{4}{*}	{Ionosphere}&	Without-oversampling&	98.99\% (99.28\%, 0.1505)&	92.53\% (94.36\%, 1.1590)&	0.9238 (0.9425, 0.01184)&	0.9289 (0.9483, 0.0122)&	0.9253 (0.9436, 0.0115)\\
	&SMOTE&	98.63\% (98.88\%, 0.0878)&	91.12\% (92.95\%, 1.159)&	0.9090 (0.9276, 0.0117)&	0.9164 (0.9366, 0.0130)&	0.9112 (0.9295, 0.0115)\\
	&GBO&\textbf{99.31\%} (99.44\%, 0.1464)&	93.09\% (95.77\%, 1.6862)&	0.9293 (0.9571, 0.0174)&	0.9360 (0.9603, 0.0162)&	0.9309 (0.9577, 0.0168)\\

	&SSG&	99.17\% (99.26\%, 0.0316)&\textbf{93.24\%} (94.36\%, 1.11)&	\textbf{0.9314} (0.9431,0.0115)&\textbf{0.9339} (0.9442, 0.0101)&\textbf{0.9323} (0.9436, 0.0111)\\\hline
\multirow{4}{*}{Spambase}&	Without-oversampling&	94.77 (94.99\%, 0.1349)&	93.08\% (93.58\%, 0.3328)&	0.9307 (0.9358, 0.0033)&	0.9308 (0.9358, 0.0033)&

	0.9308 (0.9358, 0.0033)\\

	&SMOTE&	94.05\% (94.33\%, 0.2342)&	92.38\% (92.71\%, 0.2584)&	0.9240 (0.9273, 0.0025)&	0.9248 (0.9282,0.0023)&	0.9238 (0.9272, 0.0026)\\
	&GBO&	94.92\% (95.33\%, 0.2217)&	92.32\% (93.15\%, 0.4132)&	0.9231
(0.9315, 0.0041)&	0.9232 (0.9316, 0.0029)&	0.9231 (0.9315, 0.0038)\\
&SSG&\textbf{95.72\%} (95.99\%, 0.1546)&\textbf{93.21\%} (93.58\%, 0.2907)&
	\textbf{0.9320} (0.9357, 0.0029)&\textbf{0.9322} (0.9358, 0.0029)&\textbf{0.9321} (0.9358, 0.0029)\\\hline
\multirow{4}{*}{Shuttle}	&Without-oversampling&	99.83\% (99.92\%, 0.0896)&	99.84\% (99.93\%, 0.0738)&	0.9983 (0.9993, 0.0010)&	0.9982 (0.9993, 0.0014)&	0.9987 (1.0, 0.0007)\\
	&SMOTE&	99.93\% (99.94\%, 0.0070)&	99.85\% (99.88\%, 0.0190)&	0.9984 (0.9989, 0.0009)&	0.9980 (0.9991, 0.0031)&	0.9985 (0.9989, 0.0001)\\
	&GBO&	99.93\% (99.97\%, 0.0463)&	99.85 (99.93\%, 0.0729)&	0.9989 (0.9989, 0.0001)&	0.9982 (0.9993, 0.0014)&
	0.9988 (1.0, 0.0007)\\
		&SSG&\textbf{99.95\%} (99.97\%, 0.0130)&\textbf{99.91\%} (99.95\%, 0.0327)&\textbf{0.9991} (0.9995, 0.0004)&\textbf{0.9991} (0.9995, 0.0003)&\textbf{0.9991} (0.9995, 0.0003)\\\bottomrule

    \end{tabular}}
    
    \label{tab:result}
\end{table*}

To understand the classification effect, the overall misclassification is presented in Table~\ref{tab:msclass}, where the total number of false positives and false negatives predicted by each oversampling methods are counted for each dataset. Here, the algorithm with the lowest misclassification is highlighted with bold fonts and is the primary concern that will help identify the true potential of the algorithm's performance on imbalanced datasets. From the table, it can be observed that SSG demonstrates better performance on Page blocks, Ecoli, Winequality, Abalone, Ionosphere, Spambase, Shuttle, Yeast by misclassifying 1 (0.21\%), 2(0.59\%), 6 (0.92\%), 80 (1.92\%), 6 (1.7\%), 62 (1.34\%), 11(0.02\%), and 6 (1.17\%) samples respectively. GBO becomes the second-best algorithm by misclassifying 1(0.21\%), 1 (0.084\%), and 8 (1.56\%) samples on Page blocks and yeast datasets, respectively. On the other hand, worst performance was observed when the experiment was carried out on raw dataset without Oversampling approaches. 

\begin{table}[!ht]
\caption{Misclassification occurred by different oversampling techniques used in various datasets.}
    \centering
    \begin{tabular}{ccccc}\toprule
    \multirow{2}{*}{Dataset}& \multicolumn{4}{c}{Misclassification = False positive ($F_p$) + False negative ($F_n$)}\\\cmidrule{2-5}
    &Without sampling&	SMOTE&	GBO&	SSG\\\midrule
Pageblocks&	6&	2&	\textbf{1}&	\textbf{1}\\
Ecoli&	5&	3&	3&	\textbf{2}\\

Winequality&	10&	13&	7&	\textbf{6}\\
Abalone&	122&	122&	85&	\textbf{80}\\
Ionsophere&	18&	8&	7&	\textbf{6}\\
Spambase&	85&	77&	73&\textbf{62}\\
Shuttle&	39&	20&	40&	\textbf{11}\\
yeast&	12&	9&	8&	\textbf{6}\\\bottomrule

    \end{tabular}
    
    \label{tab:msclass}
\end{table}
The optimal results revealed that the proposed SSG shows higher accuracy, precision, recall, and F1 score in most of the highly imbalanced benchmark datasets used in this study. However, during the experiment, it was noticed that even though the accuracy result between the proposed oversampling techniques and SMOTE techniques may not significantly differ, the benefit of the proposed model can be easily identified once the total misclassification is calculated (as shown in Table~\ref{tab:msclass}).

Apart from this, the proposed SSG model is the second-best algorithm (in terms of training accuracy) on Pageblocks, Yeast, and Ionosphere datasets, demonstrating the necessity of tuning the parameters of the SSG model for individual datasets instead of relying on constant parameter settings. However, in this study, the whole experiment was carried out with the fixed-parameter settings to properly compare the performance of the proposed model on referenced datasets. Additionally, the proposed GBO algorithms show a better classification effect than without sampling and SMOTE on all datasets regarding total misclassification rate.

\begin{figure*}
    \centering
    \includegraphics[width=\textwidth]{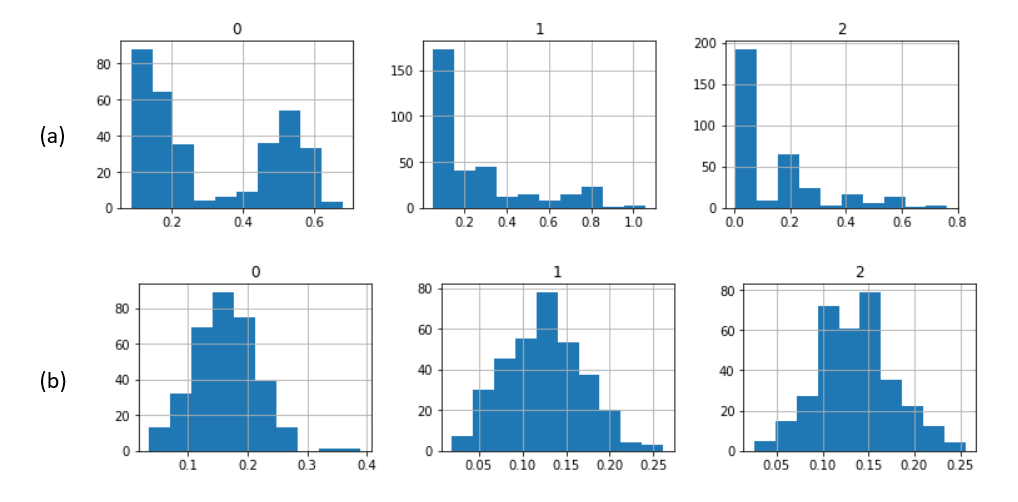}
    \caption{Performance observation using histogram of attribute of Abalone dataset using (a) SMOTE and (b) SSG.}
    \label{fig:distrib}
\end{figure*}

Figure~\ref{fig:distrib} displays the oversampling effect on the Abalone dataset. The figure shows that the sample created using the proposed SSG (Figure~\ref{fig:distrib}b) follows better Gaussian distribution than the sample generated by the original SMOTE (see Figure~\ref{fig:distrib}a).

The experimental results for standard deviation are displayed in Table~\ref{tab:interclass}, with the best experimental results in bold font.
According to Table~\ref{tab:interclass}, when the proposed SSG is used, the inter-class distance between categories for Ecoli, Winequality, Abalone, Ionosphere, and shuttle dataset is closest to the raw/original data. On the other hand, when the GBO approach is used on the Pageblocks, Spambase, and Yeast datasets, the interclass distance across categories after Oversampling is closest to the original data.
\begin{table}[!ht]
 \caption{Effect of SMOTE and proposed sampling techniques on the imbalanced dataset in terms of standard deviation.}
    \centering
    \begin{tabular}{ccccc}\toprule
    \multirow{2}{*}{Dataset}&\multicolumn{4}{c}{Standard Deviation}\\\cmidrule{2-5}	
    &Without oversampling&	SMOTE&	GBO&	SSG\\\midrule
    Pageblocks&	0.1782&	0.1927&	\textbf{0.1630}&
	0.1578\\
Ecoli&	0.1653&	0.1141&	0.0737&	 \textbf{0.2069}\\
Winequality&	0.1549&	0.1268&	0.1872&	\textbf{0.1607}\\
Abalone&	0.2631&	0.2606&	0.1681&	\textbf{0.2628}\\
Ionosphere&	0.4710&	0.4036&	0.4498&
	\textbf{0.4545}\\
	Spamebase&	0.0544&
	0.05558&
	\textbf{0.0547}&	0.05041\\

Shuttle&	44.6300&	36.8300&	47.3400&	\textbf{41.3200}\\

Yeast&	0.0915&	0.0655&	\textbf{0.0723}&	0.1677\\\bottomrule
    \end{tabular}
   
    \label{tab:interclass}
\end{table}

To verify the overall study findings, future works should apply several ML approaches such as random forest, decision tree, and XGB-Boost on similar benchmark datasets that are highly imbalanced and compare the study results with the proposed GBO and SSG algorithms outcomes.
\section{Conclusion}
The study aims to address the classification problem of the imbalanced dataset and the limitations of the existing SMOTE approaches by proposing two new Oversampling techniques based on the generative adversarial network (GAN). The study findings suggest that the effect of the proposed models demonstrates better Gaussian distribution with less marginalization than the original SMOTE on eight high imbalanced benchmark datasets. The preliminary computation results show that the proposed GAN-based Oversampling (GBO) and SVM-SMOTE-GAN (SSG) can produce synthetic samples. A neural network can achieve a higher classification effect in terms of accuracy, precision, recall, and F1 score. The misclassification rate by the proposed algorithm–GBO and SSG–is lower than the original SMOTE on benchmark datasets used in this study (refer to Table~\ref{tab:interclass}). Future studies should apply these methods in other datasets, apply mixed-data analysis using kernel methods and combine the proposed approaches with convolutional neural network (CNN) or recurrent neural network (RNN) to deal with the time series data. Further, it would be an exciting opportunity to investigate the combination of GAN with other popular Oversampling techniques such as borderline SMOTE and ADASYN.

\bibliographystyle{unsrt}  
\bibliography{main}

\end{document}